\documentclass[sigconf]{acmart}

\usepackage{booktabs} 

\begin{document}
\title{Crowd-Labeling Fashion Reviews with Quality Control}

\author{Iurii Chernushenko, Felix A. Gers, \\ Alexander L{\"o}ser}
\affiliation{%
  \institution{Beuth University of Applied Sciences}
  \streetaddress{Luxemburger Str. 10}
  \city{Berlin}
  \state{Germany}
  \postcode{13353}
}
\email{firstname.lastname@beuth-hochschule.de}

\author{Alessandro Checco}
\affiliation{%
  \institution{University of Sheffield}
  \institution{Information School}
  \streetaddress{Western Bank}
  \city{Sheffield}
  \state{UK}
  \postcode{S10 2TN}
}
\email{a.checco@sheffield.ac.uk}

\begin{abstract}
We present a new methodology for high-quality labeling in the fashion domain with crowd workers instead of experts. We focus on the Aspect-Based Sentiment Analysis task. Our methods filter out inaccurate input from crowd workers but we preserve different worker labeling to capture the inherent high variability of the opinions. We demonstrate the quality of labeled data based on Facebook's FastText framework as a baseline.
\end{abstract}


\settopmatter{printacmref=false}
\setcopyright{none} 
\settopmatter{printacmref=false}
\renewcommand\footnotetextcopyrightpermission[1]{}
\pagestyle{plain}

\maketitle

\section{Introduction}

Today, users make purchases via websites and also communicate about their opinion there. Usually, this communication happens in the form of reviews users leave under the product description. Processing of these reviews is commonly done in the form of Aspect-Based Sentiment Analysis (ABSA) - i.e., mining and summarizing opinions from the text about specific entities and their aspects. This information can help consumers decide what to purchase and businesses to better monitor their reputation and understand the needs of the market \cite{DBLP:conf/semeval/PontikiGPAMAAZQ16}. 

\textbf{Contribution.} We describe a novel quality preserving crowd labeling approach on such reviews from a highly popular European apparel online store with more than 135 million transactions monthly. Existing datasets for the ABSA task are usually labeled by experts \cite{DBLP:conf/semeval/PontikiGPAMAAZQ16}. Labeling datasets by experts is slow, expensive and therefore often cannot be executed on a larger scale. We propose a non-expert crowdsourcing procedure to get big, high-quality labeled datasets with a low budget. Our approach is based on: a) Labeling fashion text reviews accompanied by the item images. This helps to solve disambiguation problems. b) Class labeling justification. We ask crowd workers to highlight relevant text fragment, that is triggering classification. This additionally forces them to reflect on the assigned label. c) Quality assurance with a gold standard, that is filtering out inaccurate crowd workers.

\section{Fashion Review Dataset}
The initial dataset consists of 2.3 Mio. textual reviews in eleven languages written by users from the fashion store website. Each review consists of caption and text describing and/or rating the product (Figure 1). The text often has poor sentence separation, typos or unconventional punctuation usage. These problems make it harder to mine information out of the reviews.

\begin{figure}
\fbox{\includegraphics[scale=0.15]{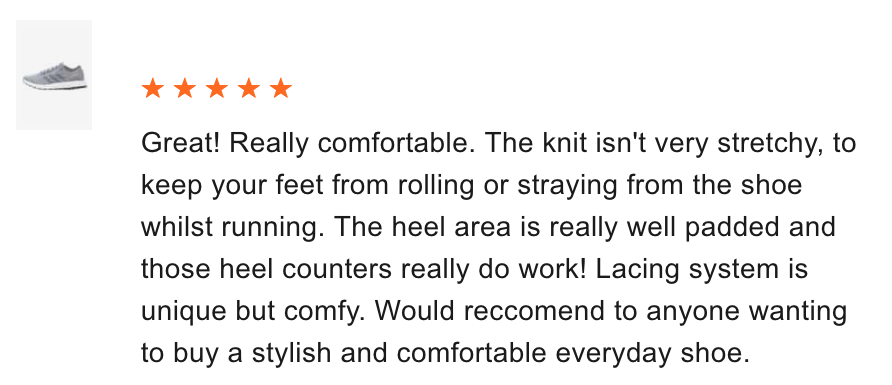}}
\caption{User review from the initial dataset.}
\end{figure}

\textbf{Aspect-selection with experts and LDA.} After processing reviews with Latent Dirichlet Allocation (LDA) \cite{DBLP:journals/jmlr/BleiNJ03} we verified potential topics with two domain experts. As a result, we unravel five common and relevant topics: General (``Perfect'', ``Fabulous!'', ``very nice''), sizing or fitting (``Wrong sizes'', ``way too big''), quality (``The quality is good and the stitching is even so the hems match and therefore is no twisting.''), design (``looks fabulous with a lurex hat''), delivery (``Quick delivery. Great!'').

\textbf{Focus on sizing issues for shoes.} Online stores in the fashion industry desire presenting accurate information about the size of their items. However, products from different manufacturers are not consistent in sizing information. Reviews of customers regarding sizing aspect differ depending on the product category. Here, we focus on labeling in the footwear category for the sizing aspect. For 3759 reviews related to shoes, we classify sentiments of Size/Fit opinions expressed in reviews as following classes: 1. Positive 2. Neutral 3. Negative 4. Other (not related to size/fit) 5. Data error (not a shoe, incorrect language, other data inconsistency).

\section{Quality Assurance}
\textbf{Ask one straightforward question.} First, we formulated the task and wrote instructions\footnote{Full instructions for our dataset and labeling examples at \url{https://github.com/yury-chernushenko/papers/tree/master/2018_www}} for crowd workers [2]. Initially, we tried to label among several classes (aspects of the opinions), but this causes confusion and errors in labeling. Hence, the task should be very simple for the non-experts to produce high-quality results.

\textbf{Ask for justification.} For every class we asked crowd workers to highlight corresponding spans (text fragments) in the review that trigger a correct classification. This approach enforces crowd workers to justify their labeling and leads to a better quality. 

\textbf{Image and text context.} Every text review we accompanied with an image of the product (Figure 2). This helps crowd workers to disambiguate while reasoning about the review and is particularly important for the fashion domain.

\begin{figure}
\fbox{\includegraphics[scale=0.15]{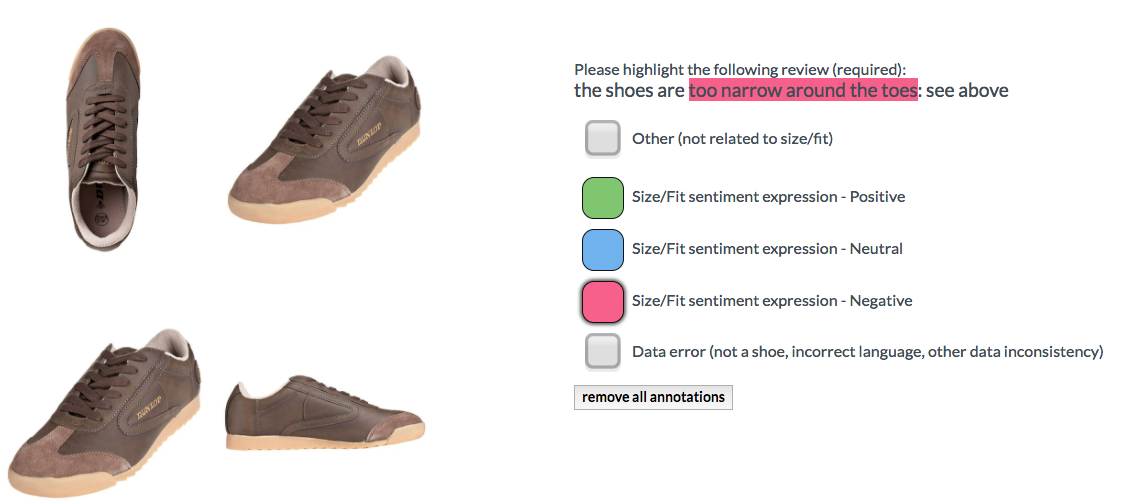}}
\caption{Labeling interface.}
\end{figure}

\textbf{Create a gold standard.} We labeled 500 reviews by two domain experts and use this data to filter out inaccurate crowd workers. We iterated on this step 3 times to create high-quality gold standard labels and to choose gold labels that are simple enough to assess general crowd worker performance. 

\textbf{Quality control.} Prelabeled gold standard questions are integrated throughout the whole labeling process. We compare the span of a crowd worker and of the gold standard by measuring the overlap relative to the size of the corresponding sentences. This approach is similar to the Intersection-over-Union (IoU) method proposed for object detection on image datasets \cite{DBLP:journals/ijcv/EveringhamEGWWZ15}. We compare two spans by limiting minimum required intersection/union ratio (0.7 in our case) with corresponding gold standard span. If crowd worker is having low error rate on the gold standard questions, then we keep other labels from this crowd worker. 

\textbf{Keep high quality while running at scale.} Crowd workers receive instructions and test questions in the beginning. Every crowd worker labels not more than 300 reviews, to avoid bias or worker attention degradation. Every item is labeled 3 times by different crowd workers to absorb diverse opinions. 

\section{Classification Baseline}
\textbf{The final label distribution} for our dataset is: Other - 6.323, Positive - 2.194, Neutral - 206, Negative - 3.574. According to this class distribution, classification precision should be higher than 0.51 (just predicting class ``Other''). We created our baseline with a popular classification library called FastText \cite{DBLP:conf/eacl/GraveMJB17}. FastText uses a bag of words approach and subword information. This functionality is beneficial for reviews with typos and mistakes. As preprocessing step we lowercase the text and remove punctuation. The neutral class is disproportional to others because customers usually express an extreme sentiment in the reviews. To express this imbalance we present weighted average of metrics. After an informed parameter grid search, we configured FastText with Wikipedia pre-trained word embeddings, 25 epochs, 3 word n-grams and set all other parameters to default. Table 1 reports our results after splitting 80/20 on train/test.

\begin{table}[]
\centering
\caption{Reviews classification results.}
\begin{tabular}{|l|l|l|l|}
\hline
\textbf{Class}     & \textbf{Precision} & \textbf{Recall} & \textbf{F1} \\ \hline
Other              & 0.91               & 0.95            & 0.93        \\ \hline
Positive           & 0.86               & 0.90            & 0.88        \\ \hline
Neutral            & 0.67               & 0.19            & 0.29        \\ \hline
Negative           & 0.91               & 0.84            & 0.88        \\ \hline
Average (weighted) & 0.90               & 0.90            & 0.90        \\ \hline
\end{tabular}
\end{table}

Please note, we also observe similar results using pre-trained word embeddings from the amazon reviews dataset of  ``Clothing, Shoes and Jewelry''. 

Table 2 shows classification results of spans only. Instead of taking whole reviews, we use only spans of highlighted text by crowd workers with corresponding classes. We consider text that is not part of any span as part of class ``Other''.

\textbf{Understanding errors.} Reviews, where the algorithm is failing, are on average 30\% longer. Third of these errors is coursing difficulty to distinguish Size/Fit and Comfortability aspects. This problem could be addressed by choosing the more specific definition of the aspect. Another 20\% of errors are caused by the lack of consensus among crowd workers. In this case, the majority of crowd workers are labeling some review correctly, but a minor incorrect label was assessed during the test run. This problem could be addressed with an aggregation by review and assigning majority voted label.

\begin{table}[]
\centering
\caption{Spans classification results.}
\begin{tabular}{|l|l|l|l|}
\hline
\textbf{Class}     & \textbf{Precision} & \textbf{Recall} & \textbf{F1} \\ \hline
Other              & 0.98               & 0.99            & 0.99        \\ \hline
Positive           & 0.97               & 0.93            & 0.95        \\ \hline
Neutral            & 0.27               & 0.23            & 0.25        \\ \hline
Negative           & 0.94               & 0.92            & 0.93        \\ \hline
Average (weighted) & 0.97               & 0.97            & 0.97        \\ \hline
\end{tabular}
\end{table}

\section{Summary and Future Work}
\textbf{Labeling quality assurance ensures strong baseline.} Our results indicate that labeling for Aspect-Based Sentiment Analysis can be done with a crowdsourcing approach at a large scale. The combination of different measures yields in high-quality data. Restricting the input to the text spans containing the opinion target expressions further improved the classification results. This indicates that labeled spans contain the most significant information. 

\textbf{Future work.} Generally, there are two types of mislabeling for a specific review: disagreement on class between crowd workers and disagreement on span content. As future work, we plan to leverage on Opinion Term Expressions for better classification results and will apply neural networks with attention mechanism.

\section{Acknowledgements}
Our work is funded by the European Union's Horizon 2020 research and innovation programme under grant agreement No 732328 (FashionBrain).

\bibliographystyle{ACM-Reference-Format}
\bibliography{paper}

\end{document}